\documentclass[sigconf,nonacm]{acmart}

\AtBeginDocument{%
  \providecommand\BibTeX{{%
    \normalfont B\kern-0.5em{\scshape i\kern-0.25em b}\kern-0.8em\TeX}}}

\usepackage{graphicx}
\usepackage{subcaption}
\usepackage{pifont}
\usepackage{enumitem}
\usepackage{algorithm}
\usepackage[noend]{algpseudocode}
\usepackage{multirow}
\usepackage{makecell}
\usepackage{booktabs}

\begin{document}

\title{GeoEdit: Adversarial De-Evidencing for Silent Geoprivacy Protection via Image Editing}

\author{Yu Yan\textsuperscript{1}, Shengjia Cheng\textsuperscript{1}, Fei Ma\textsuperscript{2}, Zhe Wang\textsuperscript{1}, and Peng Chen\textsuperscript{1}}
\affiliation{%
  \institution{\textsuperscript{1}People's Public Security University of China}
  \city{Beijing}
  \country{China}}
\affiliation{%
  \institution{\textsuperscript{2}Guangming Laboratory}
  \city{Shenzhen}
  \country{China}}
\email{yanyu7560@163.com, chenpeng@ppsuc.edu.cn}
\thanks{Corresponding author: Peng Chen, chenpeng@ppsuc.edu.cn}

\renewcommand{\shortauthors}{Yu Yan et al.}

\begin{abstract}
Recent efforts have demonstrated the strong capabilities of Multimodal Large Reasoning Models (MLRMs) in geographic reasoning. By jointly analyzing subtle visual cues, these models can infer the locations of photos, thus enabling doxing even by non-expert adversaries. 
Existing defense methods struggle to balance the shareability of photos with effective protection, and often fail to sufficiently disrupt the implicit geo-evidence chains embedded in scene semantics. 
To this end, we propose GeoEdit, an adversarial de-evidencing framework for silent geoprivacy protection via image editing. 
Specifically, we apply minimal edits to high-risk evidence anchors in the image, thereby disrupting the underlying spatial reasoning chains while preserving overall scene semantics. We identify these anchors by combining structured knowledge with adversarially discovered geographic cues. To further navigate the large editing action space, we adopt a learnable policy to iteratively refine editing decisions, distilling successful trajectories into transferable natural-language insights for consistent and efficient editing. 
Evaluated on LifePhotoBench, GeoEdit achieves strong protection against 6 mainstream MLRMs, outperforming visual perturbation baselines by over 0.59 in image fidelity.
\end{abstract}



\maketitle

\hypersetup{pdfauthor={Yu Yan, Shengjia Cheng, Fei Ma, Zhe Wang, Peng Chen},
  pdfkeywords={geoprivacy, image editing, MLRM, adversarial defense}}
\ifdefined\pdfinfo\pdfinfo{/Author (Yu Yan, Shengjia Cheng, Fei Ma, Zhe Wang, Peng Chen)}
  \pdfinfo{/Keywords (geoprivacy, image editing, MLRM, adversarial defense)}\fi

\section{Introduction}
\label{sec:introduction}

Recent Multimodal Large Reasoning Models (MLRMs), such as GPT-5.4 and Gemini-3.1, can perform multi-step reasoning~\cite{yan2026red} and infer implicit information from visual inputs~\cite{MMLmodelsEval2025, yue2024mmmu, lu2023mathvista, Xu2025VisuLogic, li2025recognition}, thereby enabling them to systematically capture subtle, high-level cues that are not explicitly stated and achieve strong performance on complex document understanding and multi-hop visual question answering~\cite{xiang2026bridge, WebQA2022, song2025bridgegap}.

\begin{figure}[t]
  \centering
  \begin{subfigure}{\linewidth}
    \includegraphics[width=0.9\linewidth]{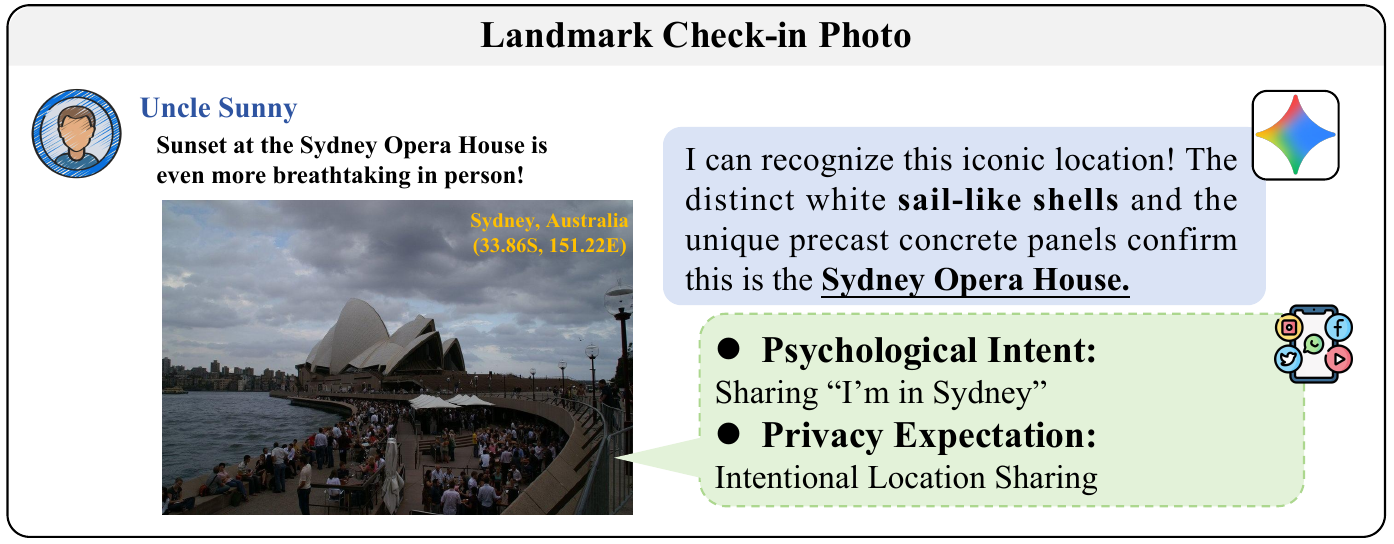}
    \caption{Landmark Check-In Photo (Intentional Location Sharing)}
    \label{fig:case_landmark}
  \end{subfigure}
  \par\vspace{1em}
  \begin{subfigure}{\linewidth}
    \includegraphics[width=0.9\linewidth]{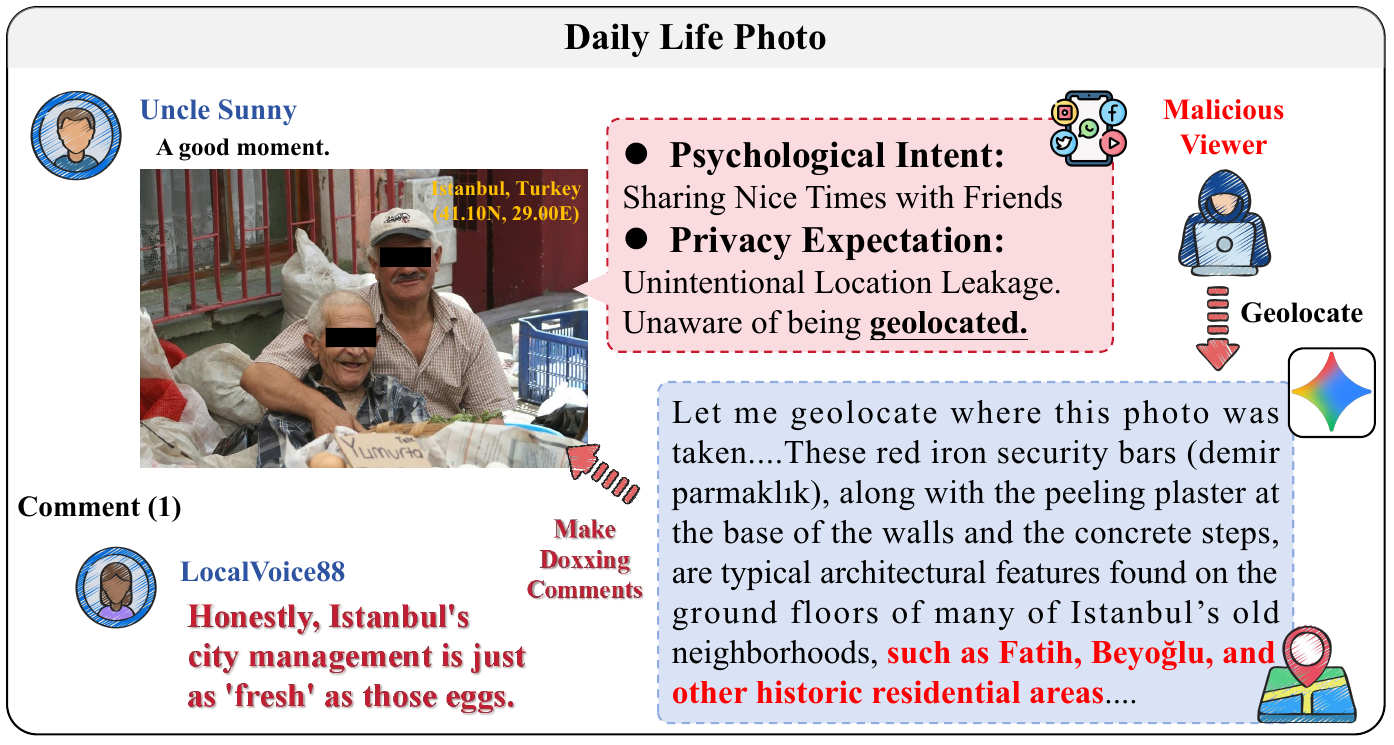}
    \caption{Daily Life Photo (Unintentional Geolocation Doxing Risk)}
    \label{fig:case_daily}
  \end{subfigure}
  \caption{Comparison of social media sharing scenarios. The landmark check-in (a) exemplifies intentional location sharing. Conversely, the daily life photo (b) inadvertently leaks sensitive location data via implicit geo-evidence, resulting in non-consensual exposure and severe doxing risks.}
  \label{fig:cases}
  \vspace{-10pt}
\end{figure}

\begin{figure*}[ht]
  \centering
  \begin{subfigure}{0.24\textwidth}
    \includegraphics[width=\linewidth]{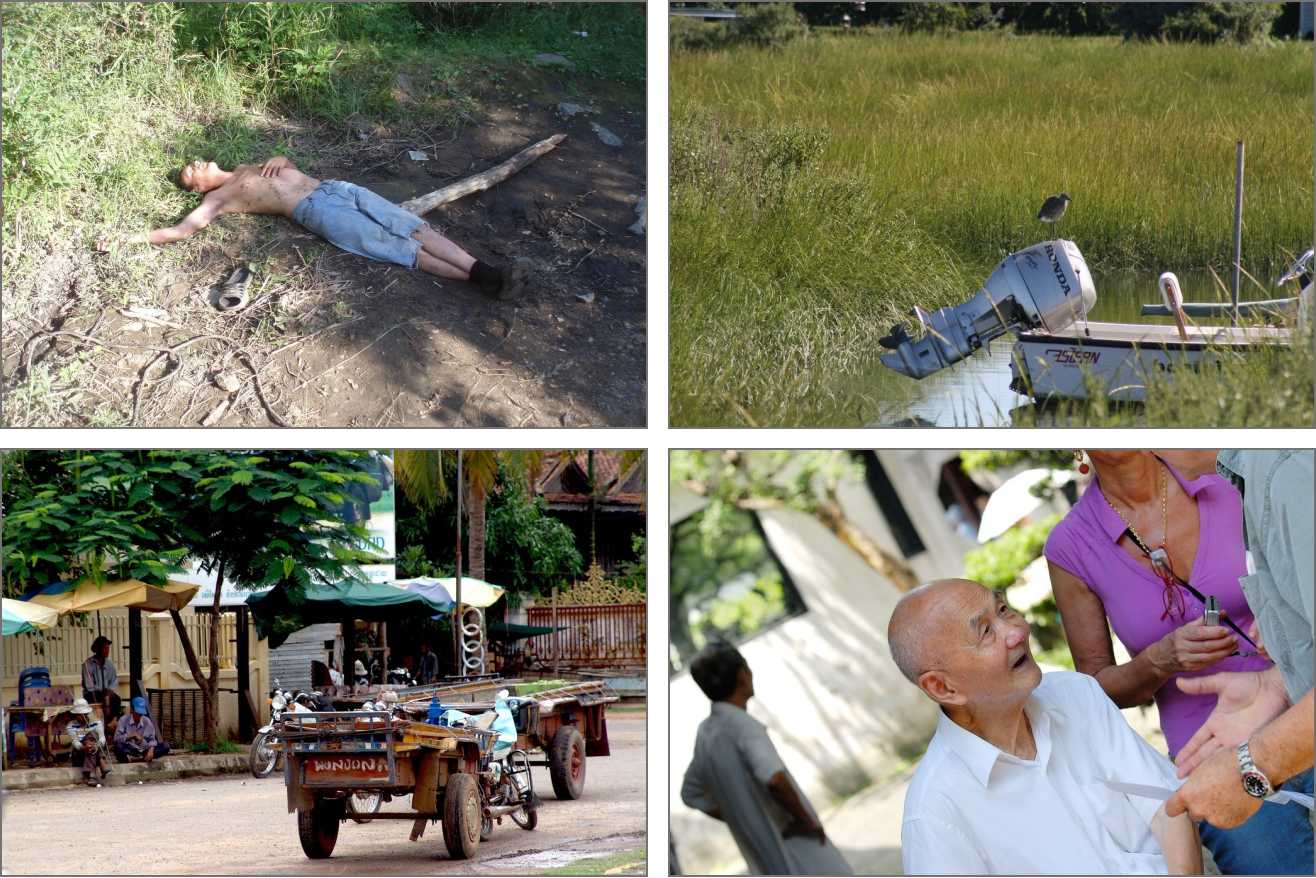}
    \caption{Original Image}
    \label{fig:method_ori}
  \end{subfigure}
  \hfill
  \begin{subfigure}{0.24\textwidth}
    \includegraphics[width=\linewidth]{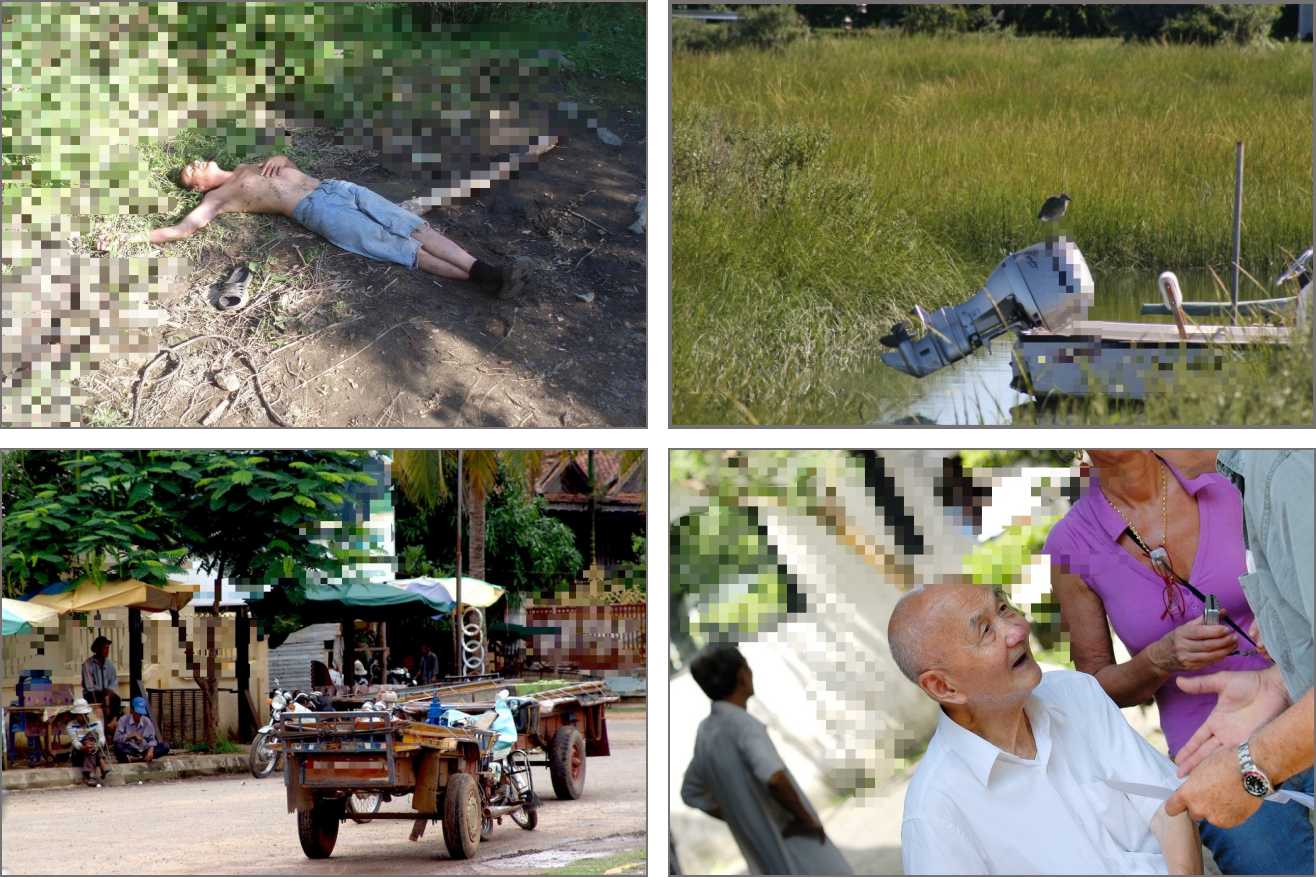}
    \caption{Image Obfuscation}
    \label{fig:method_mask}
  \end{subfigure}
  \hfill
  \begin{subfigure}{0.24\textwidth}
    \includegraphics[width=\linewidth]{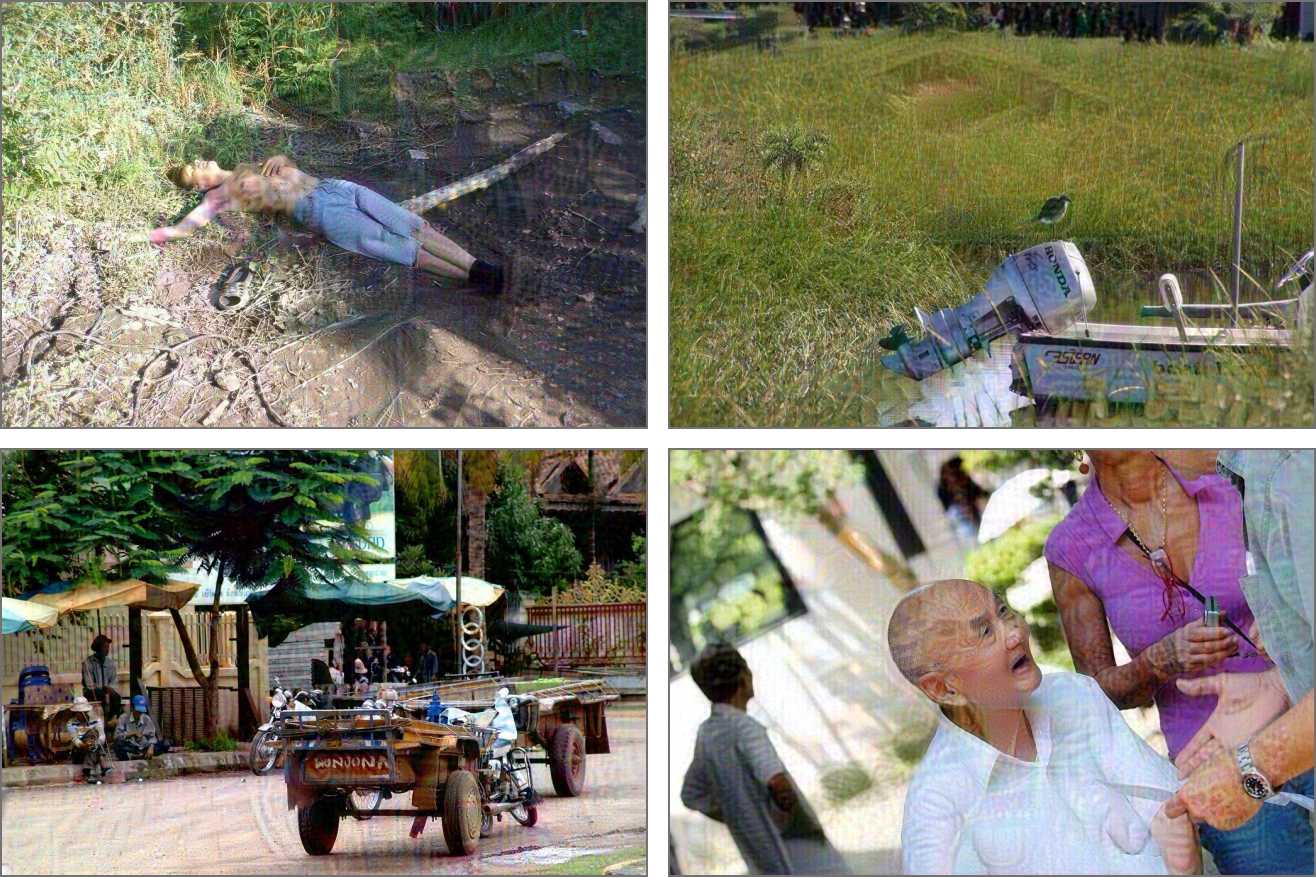}
    \caption{Visual Perturbation}
    \label{fig:method_geoshield}
  \end{subfigure}
  \hfill
  \begin{subfigure}{0.24\textwidth}
    \includegraphics[width=\linewidth]{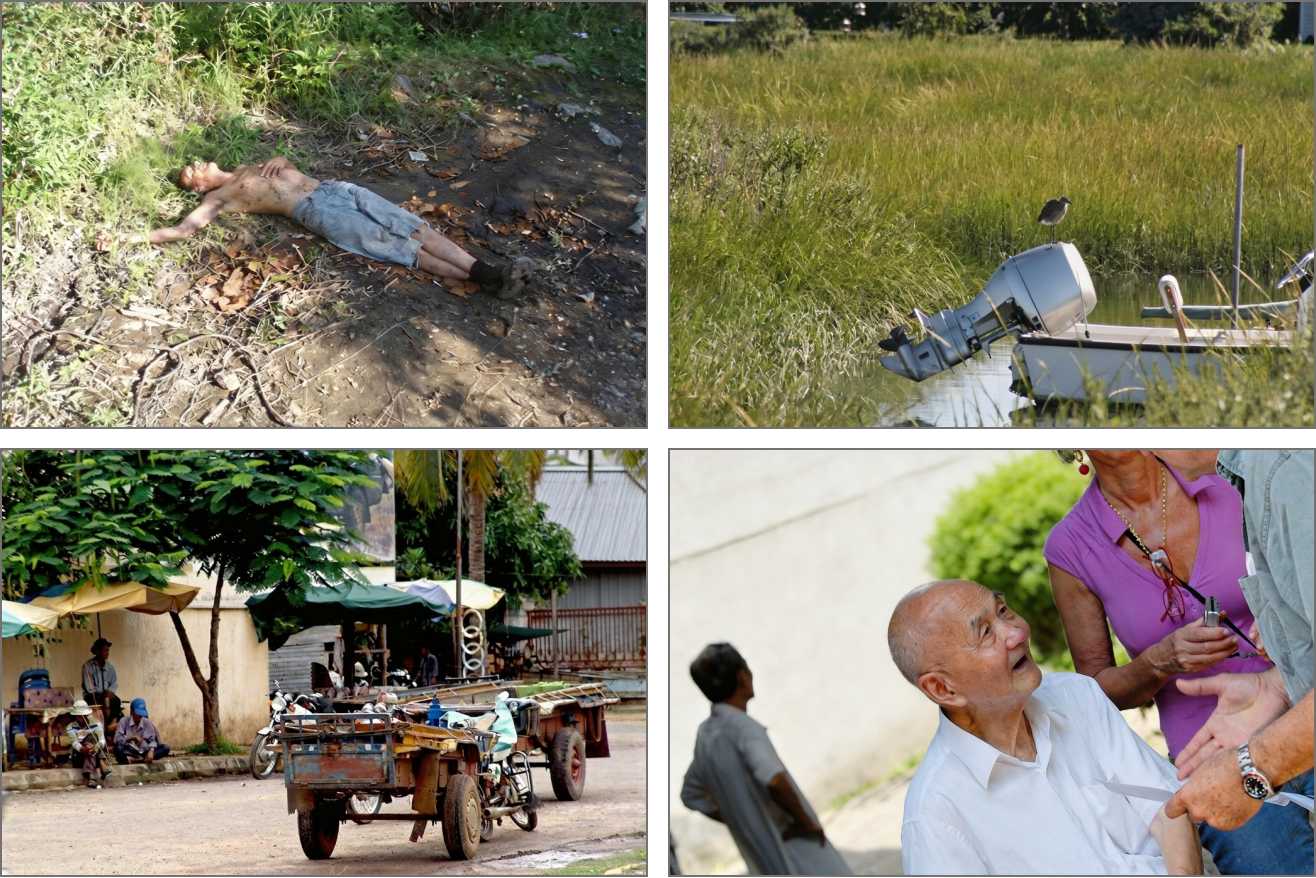}
    \caption{De-Evidencing Editing (Ours)}
    \label{fig:method_geoedit}
  \end{subfigure}
  \vspace{-10pt}
  \caption{Comparison of geoprivacy protection mechanisms. (a) The unprotected image. (b) Image Obfuscation hides sensitive areas but ruins the visual experience. (c) Visual Perturbation tricks visual encoders to protect geoprivacy, but its adversarial nature is often lost during unpredictable image processing and causes noticeable artifacts. (d) \textbf{Geo-Evidence Editing (Ours)} blocks the localization reasoning chain to defend against both MLRMs and human analysis while keeping visual appeal.}
  \label{fig:defense_comparison}
  \vspace{-5pt}
\end{figure*}

However, this broader reasoning ability introduces new privacy and security risks in our everyday image-sharing scenarios. 
As illustrated in Figure~\ref{fig:cases}, in contrast to landmark check-ins (Figure~\ref{fig:case_landmark}), where users explicitly disclose their location, daily-life photos (Figure~\ref{fig:case_daily}) typically do not include clear landmarks and instead focus on personal moments or social interactions. As a result, these images are often shared without considering the risk of location disclosure, expecting the model to respect contextual integrity~\cite{yang2026contextual}.
However, even without explicit landmarks, such images often contain multiple fragmented clues, such as architectural styles, road infrastructure, and textual context~\cite{weyand2016planet}, which can jointly reveal sensitive location information in a more implicit and less noticeable way.
In most cases, exploiting such fragmented clues requires trained Open-Source Intelligence (OSINT) analysts \cite{pastorgalindo2020osint, hassan2018osint, higgins2021bellingcat}. However, with the emergence of MLRMs, this threat landscape fundamentally changes, as joint visual-semantic reasoning enables even non-experts to infer locations from subtle geo-evidence at fine granularity \cite{openai2023gpt4v, geminiteam2023gemini, luo2025doxing, zhang2025disrupting, li2025pixels}. As a result, precise geolocation doxing is increasingly automated and scalable~\cite{zhang2025evaluation}.

Some efforts have been made to develop geoprivacy protection methods to address this threat, as illustrated in Figure~\ref{fig:defense_comparison}. However, their limitations remain in practice, particularly when balancing protection strength with visual utility. Explicit image obfuscation methods, such as cropping, blurring, and masking (Figure~\ref{fig:method_mask}), can conceal visible cues~\cite{orekondy2017visual, shu2016cardea, orekondy2018connecting}, but they often compromise semantic completeness and visual naturalness, while also leaving clear signs of protection that discourage users from sharing images.
Recent methods introduce visual adversarial perturbations (Figure~\ref{fig:method_geoshield}) to add noise that reduces the localization performance of multimodal models~\cite{zhang2020adversarial, guo2024efficient, li2025frustratingly, liu2025geoshield}. However, these methods mainly affect low-level perception rather than removing geo-semantic evidence, and the underlying evidence chain remains largely unchanged~\cite{zhang2025disrupting}. As a result, both human OSINT analysis and advanced models can still recover location information by combining the remaining cues. Moreover, such perturbations are often model-specific and tend to overfit, making them fragile under model evolution or common image transformations in real-world settings.


To this end, we aim to achieve more generalizable geoprivacy protection by introducing targeted, localized modifications that weaken relations among geo-evidence cues, thereby disrupting the spatial reasoning chain.
We propose \underline{\textbf{Geo}}-Evidence \underline{\textbf{Edit}}ing (\textbf{GeoEdit}), a de-evidencing image editing framework for \textbf{silent geoprivacy protection}. GeoEdit combines structured knowledge with adversarially discovered geo cues to identify high-risk evidence anchors and applies minimal edits to break their mutual corroboration while preserving scene semantics and visual naturalness.
This de-evidencing process is inherently challenging, as the editing space is large, subtle, and structured around geo-evidence cues. Navigating this space requires identifying the visual elements that support geolocation, understanding how they jointly contribute to geographic reasoning, and modifying them in a restrained manner that avoids visible artifacts.
To address this, GeoEdit introduces a test-time insight distillation mechanism built on an agentic process that includes evidence discovery, edit synthesis, and adversarial verification. From this process, the framework extracts successful editing trajectories and summarizes them as transferable natural-language protection insights to guide future edits and improve cross-sample robustness. Extensive experiments on LifePhotoBench show that GeoEdit achieves state-of-the-art protection against 6 mainstream MLRMs. 
It neutralizes precise geolocation accuracy (Acc@1km) to near 0\% across all MLRMs, and can achieve a remarkable Median Distance Error (MDE) of over 5000\,km on existing mainstream MLRMs. Meanwhile, it maintains a high fidelity of 0.944, far surpassing existing visual perturbation methods (typically below 0.35).
Our major contributions are as follows:

\begin{itemize}[leftmargin=1em]
\item We propose \textbf{GeoEdit}, an image-editing framework for Silent Geoprivacy Protection that disrupts geo-evidence corroboration instead of relying on obfuscation or perturbations, thereby reducing geolocation risk while preserving visual quality.
\item To navigate the large and subtle editing space structured around geo-evidence cues, we introduce a self-evolution mechanism to distill successful rollouts into transferable and interpretable natural-language protection insights.
\item Extensive experiments show that GeoEdit achieves strong protection against 6 mainstream MLRMs. It can reduce geolocation Acc@1km to near 0\% and achieves a Median Distance Error exceeding 5,000km, while maintaining high visual fidelity (0.944), surpassing perturbation-based methods by over 0.59.
\end{itemize}

\begin{figure*}[t]
  \centering
  \includegraphics[width=1.0\linewidth]{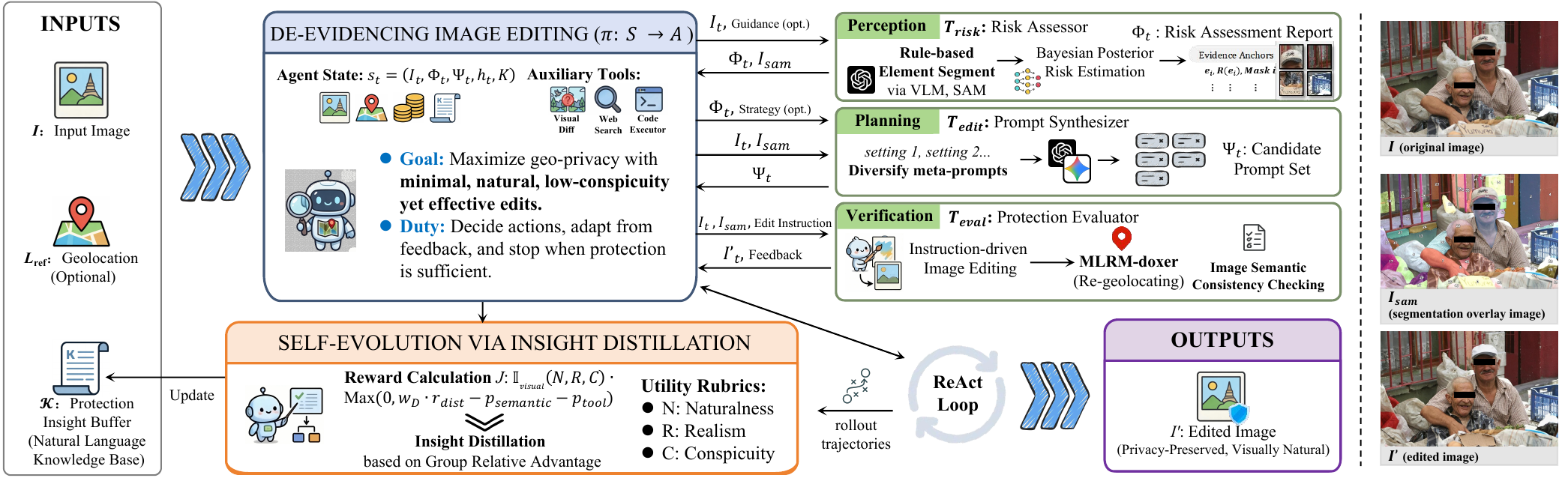}
    \caption{Overall architecture of GeoEdit. \ding{182} {De-Evidencing Image Editing}: A policy $\pi$ explicitly reasons and acts within a ReAct loop, invoking the core tools $T_{risk}$, $T_{edit}$, and $T_{eval}$ to identify evidence, generate semantic edits, and verify protection quality. 
    \ding{183} {Self-Evolution via Insight Distillation}: A test-time optimization mechanism that constructs group-relative joint reward signals from rollout trajectories, distilling them into natural-language protection insights ($\mathcal{K}$) to guide future edits.}
  \label{fig:overall}
\end{figure*}

\section{Related Work}
\label{sec:related_work}

\paragraph{\textbf{Geoprivacy Threats from Multimodal Models}}

Geoprivacy risks have broadened with the rapid advancement of multimodal models, evolving from early encoders (e.g., GeoCLIP) to current MLRMs (e.g., Gemini-2.5) with strong geolocation capabilities.

Early methods such as DeepGeo \cite{suresh2018deepgeo}, efficient feature extractors \cite{berton2022rethinking}, and multimodal alignment models like GeoCLIP \cite{vivanco2023geoclip} estimate coarse geographic locations via image retrieval or joint embedding spaces, extending privacy risks from isolated landmarks to regional identification. 
Recently, MLRMs have greatly amplified doxing threats by enabling fine-grained geographic inference from previously unexploitable daily-life photos, fundamentally transforming geo-localization into a reasoning-driven task~\cite{li2025recognition, zhang2025evaluation}. MLRM-driven methods such as PIGEON \cite{haas2024pigeon}, LLMGeo \cite{wang2024llmgeo}, GeoLocator \cite{yang2024geolocator}, and Doxing via the Lens \cite{luo2025doxing} leverage vast world knowledge, spatial graphs, and Chain-of-Thought (CoT) reasoning to synthesize subtle visual clues (e.g., architecture, vegetation, language context) \cite{li2025pixels}. 

\paragraph{\textbf{Geoprivacy Protection Mechanisms}}

To mitigate the growing risks of geoprivacy leakage, various protection mechanisms have been proposed. We categorize existing methods into two main strategies as shown in Figure \ref{fig:defense_comparison}: 
\ding{182} \textbf{Image Obfuscation.} Masking, blurring, and large-area occlusion \cite{orekondy2017visual, shu2016cardea, orekondy2018connecting} are typical methods for concealing sensitive content. While effective at blocking explicit attributes, these methods inevitably damage the image's semantic integrity and visual naturalness. The obvious tampering traces significantly degrade the user experience and reduce the image's shareability.
\ding{183} \textbf{Visual Perturbation.} Visual perturbation formulates the problem as an adversarial attack against MLRMs, injecting imperceptible noise to degrade localization performance \cite{zhang2020adversarial, guo2024efficient, li2025frustratingly, liu2025geoshield}. However, this approach misaligns with the actual localization mechanisms of MLRMs and OSINT experts, as it only disrupts model perception without removing the underlying geographic semantics. Furthermore, pixel-level perturbations are highly fragile and easily neutralized by routine image compression or resizing.



\section{Methodology}
\label{sec:method}

We design GeoEdit as an evidence-level editing framework for geoprivacy protection, which focuses on identifying and modifying key location-sensitive cues rather than introducing fragile and visually unnatural perturbations. 
We highlight our {trap-like} \textbf{Silent Geoprivacy Protection} strategy, in which the defense preserves visual naturalness under normal viewing while covertly interfering with geolocation inference when it is attempted.

\subsection{Architecture Overview}

As shown in Figure~\ref{fig:overall}, GeoEdit consists of \textbf{De-Evidencing Image Editing} and \textbf{Self-Evolution via Geoprivacy Protection
Insight Distillation}. Specifically, \ding{182} The image editing process follows a ReAct loop \cite{yao2022react}, enabling the policy $\pi$ to explicitly reason about location-sensitive evidence, propose edits, and verify whether an edit improves geoprivacy while preserving visual quality for the current sample. 
\ding{183} A self-evolution mechanism is further introduced to address the short-sightedness of per-sample decisions, thereby avoiding overly aggressive edits, unstable preferences across rollouts, and limited interpretability of the edits.

\subsection{De-Evidencing Image Editing}
The core of GeoEdit is an agentic protection policy $\pi: \mathcal{S} \rightarrow \mathcal{A}$ that maps states to actions, where $\mathcal{S}$ is the state space and $\mathcal{A}$ is the action space. This editing space is large, subtle, and structured around geo-evidence cues. Formally, given an image $I$ and its reference location $L_{ref}$, the state at step $t$ is defined as:
\begin{equation}
s_t = (I_t, \Phi_t, \Psi_t, h_t, \mathcal{K}),
\end{equation}
where $I_t$ is the current image, $\Phi_t$ is the risk assessment report, $\Psi_t$ is the candidate prompt set, $h_t$ is the action-observation history, and $\mathcal{K}$ is the protection insight buffer. The buffer stores natural-language protection insights distilled from previous samples, which are prepended to the prompt at inference time as a cross-sample prior. $L_{ref}$ can be the ground truth or an estimated pseudo-ground-truth, and it does not rely on users providing GPS coordinates to ensure the subsequent operation. GeoEdit's main goal is to disrupt the visual evidence chain supporting geolocation, and even with an estimated $L_{ref}$, we can still introduce uncertainty, thereby increasing the difficulty for potential attackers.


To align the policy with the perception-planning-verification process of the ReAct paradigm~\cite{yao2022react}, we group low-level atomic operations into a set of unified actions. The resulting action space consists of three core tool calls, one auxiliary action, and one termination action:
\begin{equation}
A_t = \{a_t^{risk}, a_t^{edit}, a_t^{eval}, a_t^{aux}, a_t^{stop}\},
\end{equation}
where $a_t^{risk}$ captures privacy threat perception via $T_{risk}$, $a_t^{edit}$ performs strategic image manipulation via $T_{edit}$, and $a_t^{eval}$ handles execution and verification of the selected edit via $T_{eval}$. We further include $a_t^{aux}$ to invoke auxiliary utilities for additional information, and $a_t^{stop}$ to terminate the process.

\paragraph{\textbf{Risk Assessor}}
In the perception stage, the policy invokes $T_{risk}$ to analyze the image through structured, schema-guided cue extraction and Bayesian posterior estimation. It produces a risk assessment report $\Phi_t$ that contains geo-evidence anchors and their scores, along with a segmentation overlay $I_{sam}$ that assigns mask IDs to fine-grained regions for image editing:
\begin{equation}
\Phi_t, I_{sam} = T_{risk}(I_t, guidance).
\end{equation}

We formalize the execution logic of $T_{risk}$ in Algorithm~\ref{alg:trisk}. $T_{risk}$ focuses on extracting actionable, location-sensitive evidence for downstream editing instead of exhaustively listing scene details. It combines structured cue extraction, posterior risk estimation, and region-level grounding to turn a complex image into a compact, prioritized set of evidence anchors. Specifically, it estimates the geolocation probability ($P(\textit{IsSolved}=1 \mid F_{risk})$) and quantifies each cue's risk contribution via leave-one-out attribution. We integrate the Bayesian network used for this estimation from the pre-built pipeline in \citet{cheng2026beyond}. 
$T_{risk}$ also preserves the global scene context through holistic captioning, providing entity-level semantic alignment references for subsequent decision-making.

\begin{algorithm}[t]
\caption{Privacy Threat Perception via Risk Assessor}
\label{alg:trisk}
\begin{algorithmic}[1]
\State \textbf{Input:} Current state image $I_t$, optional high-level $guidance$
\State \textbf{Output:} Structured risk assessment report $\Phi_t$, segmentation overlay image $I_{sam}$
\State $F_{risk}, E \leftarrow \text{MLRM\_Extractor}(I_t, guidance)$ \Comment{Extract structured risk profile $F_{risk}$ and evidence anchors $E$}
\State $caption \leftarrow \text{MLRM\_Captioner}(I_t)$
\State $P_{total} \leftarrow P(\textit{IsSolved}=1 \mid F_{risk})$ \Comment{BN-based posterior inference}
\For{each evidence $e_i \in E$}
    \State $P_{loo} \leftarrow P(\textit{IsSolved}=1 \mid F_{risk} \setminus \{e_i\})$
    \State $R(e_i) \leftarrow P_{total} - P_{loo}$
\EndFor
\State Sort $E$ descending by $R(e_i)$ \Comment{Risk attribution via leave-one-out}
\State $I_{sam} \leftarrow \text{SAM}(I_t, E)$ \Comment{Generate segmentation overlay $I_{sam}$ with mask IDs}
\State $\Phi_t \leftarrow \{E, R(E), caption, P_{total}\}$
\State \textbf{return} $\Phi_t, I_{sam}$
\end{algorithmic}
\end{algorithm}
\vspace{-5pt}

\begin{table}[h]
\centering
\caption{The geolocation evidence schema~\cite{cheng2026beyond}.}
\label{tab:bn_schema}
\vspace{-10pt}
\resizebox{\linewidth}{!}{
\begin{tabular}{@{}ll@{}}
\toprule
\textbf{Evidence Source} & \textbf{Specific Features} \\
\midrule
Continuous Geo-Features & Similarity Mean, Similarity Variance \\
Discrete Geo-Features & Scene Complexity, Landmark Level, Text Value \\
Off-Image Features & Task Attractiveness, Geo-Hint \\
Image Quality & Weather, Focal Length, View Angle, Clarity \\
\bottomrule
\end{tabular}
}\
\vspace{-8pt}
\end{table}

As illustrated in Table~\ref{tab:bn_schema}, our Risk Assessor organizes geolocation evidence into four complementary sources by following~\cite{cheng2026beyond}, thereby decomposing vulnerability into an interpretable evidence chain with structured attribution.
Specifically, \ding{182} \textbf{Continuous Geo-Features} use \textit{Similarity Mean} and \textit{Similarity Variance} to capture the statistical profile of regional appearance against a global reference, distinguishing common and distinctive landscapes. \ding{183} \textbf{Discrete Geo-Features} capture key localization anchors: \textit{Scene Complexity} measures the density of geolocation-relevant structures, while \textit{Landmark Level} and \textit{Text Value} provide high-precision signals. \ding{184} \textbf{Off-Image Features} account for UGC/OSINT metadata, where \textit{Task Attractiveness} reflects the incentive for analysis and \textit{Geo-Hint} measures explicit geographic leakage in captions or metadata. \ding{185} \textbf{Image Quality} variables, including \textit{Weather}, \textit{Focal Length}, \textit{View Angle}, and \textit{Clarity}, characterize observability conditions that affect visual utility and inference feasibility.

\paragraph{\textbf{Prompt Synthesizer}}
In the planning stage, the policy invokes $T_{edit}$ to generate diverse editing strategies conditioned on the risk assessment report $\Phi_t$ and the segmentation overlay $I_{sam}$. Instead of committing to a single trajectory, $T_{edit}$ produces a set of candidate editing plans to support subsequent selection and aggregation:
\begin{equation}
\Psi_t = T_{edit}(\Phi_t, I_{sam}, strategy, n).
\end{equation}

Specifically, $T_{edit}$ first identifies a subset of high-contributing geo-evidence from $\Phi_t$, denoted as $E_{\mathrm{tar}} = \left(\Phi_t^{(E)}\right)_{1:k}$, which serves as the primary intervention target. It then synthesizes editing instructions under different generation settings, where each setting controls the prompt configuration and model choice to induce diverse behaviors. Through iterative expansion across settings, candidate plans are accumulated until the desired number $n$ is reached.

\begin{algorithm}[t]
\caption{Verification and State Update via Protection Evaluator}
\label{alg:teval}
\begin{algorithmic}[1]
\State \textbf{Input:} Current state $I_t$, mask overlay $I_{sam}$, instruction $c_t$, reference location $L_{ref}$, historical pool $\mathcal{H}$, tolerance $\tau$
\State \textbf{Output:} Next state image $I_{t+1}$ (updated by policy)
\State \textbf{Step 1: Protection Evaluation ($T_{eval}$)}
\State $I'_t \leftarrow \text{Edit}(I_t, I_{sam}, c_t)$ 
\State $\hat{L}(I_t), \Phi_{atk}^{base} \leftarrow \text{MLRM-doxer}(I_t)$ \Comment{Baseline attack}
\State $\hat{L}(I'_t), \Phi_{atk}^{edit} \leftarrow \text{MLRM-doxer}(I'_t)$ \Comment{Attack on edited image}
\State $\mathcal{E}(I_t) \leftarrow d_H(\hat{L}(I_t), L_{ref})$ \Comment{Distance to reference}
\State $\mathcal{E}(I'_t) \leftarrow d_H(\hat{L}(I'_t), L_{ref})$
\State $\Delta\mathcal{E} \leftarrow \mathcal{E}(I'_t) - \mathcal{E}(I_t)$ \Comment{Calculate protection gain}
\State $sim_{clip} \leftarrow \text{sim}(\text{CLIP}(I_t), \text{CLIP}(I'_t))$ \Comment{Fidelity Checking}
\State \textbf{return} $(I'_t, \Delta\mathcal{E}, \Phi_{atk}^{edit}, sim_{clip})$ to the policy $\pi$
\State \textbf{Step 2: Global Optimal State Image Update}
\If{$\Delta\mathcal{E} > 0$}
    \State $\mathcal{H} \leftarrow \mathcal{H} \cup \{(I'_t,\, \Delta\mathcal{E},\, sim_{clip})\}$
    \State $sim_{max} \leftarrow \max_{(I,\, \Delta\mathcal{E},\, s) \in \mathcal{H}} \; s$
    \State $\mathcal{H}_{cand} \leftarrow \{(I,\, \Delta\mathcal{E},\, s) \in \mathcal{H} \;\mid\; s \ge sim_{max} - \tau\}$ 
    \State $(I_{t+1}, \Delta\mathcal{E}_{best}, s_{best}) \leftarrow \arg\max_{(I,\, \Delta\mathcal{E},\, s) \in \mathcal{H}_{cand}} \; \Delta\mathcal{E}$ \Comment{Update to the candidate with the largest privacy gain}
\Else
    \State $I_{t+1} \leftarrow I_t$ \Comment{Edit rejected, maintain current state}
\EndIf
\end{algorithmic}
\end{algorithm}
\vspace{-5pt}

\paragraph{\textbf{Protection Evaluator}}
In the verification stage, the policy invokes $T_{eval}$ to execute candidate edits and assess their impact:
\begin{equation}
(I'_t, \Delta\mathcal{E}, \Phi_{atk}^{edit}, sim_{clip}) = T_{eval}(I_t, I_{sam}, c_t, L_{ref}).
\end{equation}

We outline the logic of $T_{eval}$ in Algorithm~\ref{alg:teval}. At a high level, $T_{eval}$ couples semantic editing with adversarial geolocation testing to evaluate both privacy gain and visual utility.

The geolocation verification is performed by the MLRM-doxer, which employs a CoT-Doxing strategy using geolocation evidence schema (Table~\ref{tab:bn_schema}). 
To prevent reward hacking, the MLRM-doxer is isolated from the final reward phase and driven by different MLRMs.


\subsection{Self-Evolution via Geoprivacy Protection Insight Distillation}
\label{sec:reward}
To achieve natural, realistic, and inconspicuous edits, we introduce a self-evolution mechanism~\cite{deng-etal-2025-self,wu2025evolver,cai2025trainingfreegrpo} that distills successful protection attempts into transferable cross-sample insights, thereby enabling GeoEdit to refine its strategies over time and adapt across inputs.

\begin{figure}[t]
    \centering
    \includegraphics[width=0.9\linewidth]{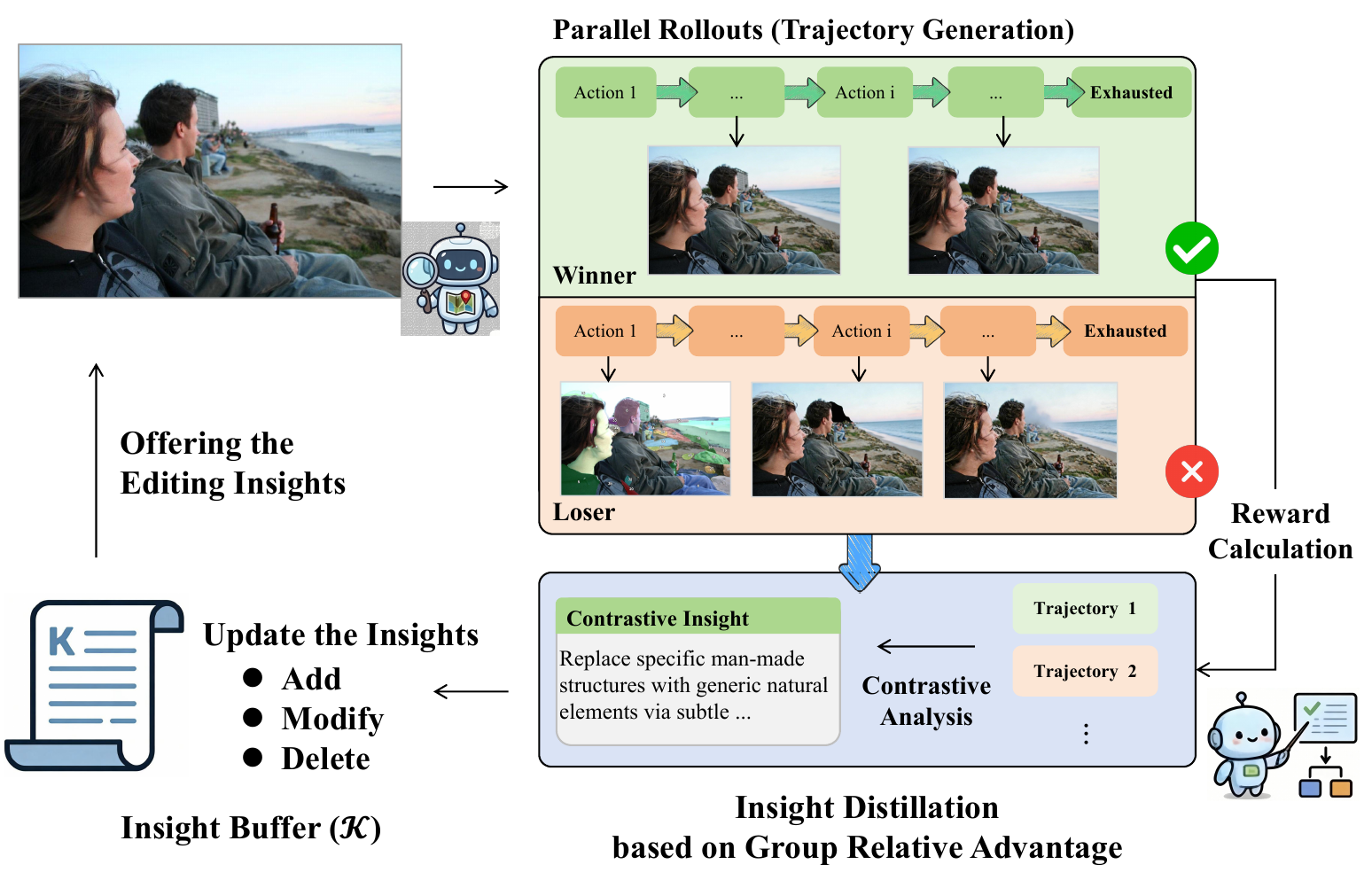}
    \caption{\textbf{Overview of Self-Evolution via Geoprivacy Protection Insight Distillation.} By contrasting successful and unsuccessful editing trajectories, GeoEdit distills a group relative advantage into a shared Insight Buffer ($\mathcal{K}$).}
    \label{fig:grpo}
\end{figure}

\paragraph{\textbf{Group Relative Advantage}}
As illustrated in Figure~\ref{fig:grpo}, insights are derived by comparing editing trajectories within each rollout group. 
For a given image, a group of editing trajectories ($G$) is generated in parallel, evaluated by a shared reward function $J(I')$, and ranked to distinguish high- and low-quality outcomes. A Judge Model then conducts contrastive analysis between high-reward (Winner) and low-reward (Loser) trajectories to distill concise natural-language insights that capture high-value actions. These insights are stored in a shared buffer $\mathcal{K}$ and injected as token-level priors during subsequent inference.

By aggregating geoprivacy protection insights across samples, the policy is guided toward more effective image changes. These insights form an explicit and interpretable set of strategies that can be directly verified by humans. They also provide an intuitive explanation of GeoEdit, showing how protection is achieved and how it can generalize to broader privacy settings.




\paragraph{\textbf{Joint Reward Formulation}}
The trajectory evaluation is defined by $J(I')$, which combines a hard constraint gate with a continuous distance reward and penalty terms:
\begin{equation}
    J(I') = \mathbb{I}_{\text{visual}}(N, R, C) \cdot \max\big(0,\ w_D\cdot r_{dist} - p_{semantic} - p_{tool}\big),
\end{equation}
where the visual utility gate $\mathbb{I}_{\text{visual}}$ applies a strict one-vote rule to maintain visual plausibility. It is assessed by a vision-language model using three discrete utility rubrics: Naturalness ($N$), Realism ($R$), and Conspicuity ($C$). Instead of aggregating these scores, the gate requires each rubric to meet its threshold, and the edit is rejected if any condition fails:
\begin{equation}
    \mathbb{I}_{\text{visual}} = \begin{cases}
      1, & \text{if } (N \ge \tau_{N}) \land (R \ge \tau_{R}) \land (C \le \tau_{C}) \land (\tau_{inv} = 0), \\
      0, & \text{otherwise},
    \end{cases}
\end{equation}
where $\tau_{N}, \tau_{R}$, and $\tau_{C}$ are tolerance thresholds for visual plausibility, and $\tau_{inv} = 1$ indicates fundamentally invalid edits such as failed inpainting or edits confined to padding regions. 

\paragraph{\textbf{Distance Reward}} The distance reward $r_{dist}$ measures protection effectiveness by jointly accounting for the achieved geolocation error (divergence from the reference location $L_{ref}$) and the improvement relative to the original image:
\begin{equation}
r_{dist} = \begin{cases}
  1.0, & \text{if } \mathcal{E}(I') \ge E_{\star}, \\[4pt]
  \alpha \cdot \mathrm{clip}\big(\frac{\mathcal{E}(I')}{E_{\star}}\big) + (1\!-\!\alpha) \cdot \mathrm{clip}\big(\frac{\max(0,\,\Delta\mathcal{E})}{D_{\star}}\big), & \text{otherwise},
\end{cases}
\end{equation}
where $E_{\star}$ is the absolute error saturation threshold, $D_{\star}$ is the gain saturation scale, and $\alpha$ controls the weight on absolute error. The reward reaches its maximum once the required protection distance is achieved, and otherwise balances absolute error and relative gain.

\paragraph{\textbf{Execution Penalties}}
The semantic penalty discourages content drift by constraining deviations between the original and edited images with penalty weight $\lambda_S$:
\begin{equation}
p_{semantic} = \lambda_S \cdot (1 - \mathrm{sim}(I, I')).
\end{equation}

The tool penalty $p_{tool}$ assigns increasing costs to repeated tool invocations, thereby encouraging efficient use of the available tools:
\begin{equation}
p_{tool} = \lambda_1 \cdot \max(0,\, n_1 - 1)
         + \lambda_2 \cdot \max(0,\, n_2 - 1)
         + \lambda_3 \cdot \max(0,\, n_3 - 1),
\end{equation}
where $n_1, n_2, n_3$ are the call counts of $T_{risk}$, $T_{edit}$, and $T_{eval}$, and $\lambda_1, \lambda_2, \lambda_3$ are the corresponding penalty weights.

\begin{table*}[t]
  \caption{Experiment results of baseline comparisons on geoprivacy protection effectiveness. Geolocation prediction accuracy (\%) is reported across different granularities. The best and second-best results are highlighted in bold and underlined, respectively.}
  \label{tab:track2}
  \centering
  \vspace{-5pt}
  \resizebox{0.96\textwidth}{!}{
  \begin{tabular}{l|c|ccccc|ccccc|ccccc}
    \toprule
    \multirow{2}{*}{\textbf{Method}} & \multirow{2}{*}{\textbf{Fid. ($\uparrow$)}} & \multicolumn{5}{c|}{\textbf{GPT-5.1 Acc@Km ($\downarrow$)}} & \multicolumn{5}{c|}{\textbf{Claude-4.5-Sonnet Acc@Km ($\downarrow$)}} & \multicolumn{5}{c}{\textbf{Claude-3.7-Sonnet Acc@Km ($\downarrow$)}} \\
    & & 1km & 25km & 200km & 750km & 2500km & 1km & 25km & 200km & 750km & 2500km & 1km & 25km & 200km & 750km & 2500km \\
    \midrule
    Unprotected & 1.000 & 29.6 & 61.9 & 71.8 & 80.3 & 83.8 & 26.5 & 55.4 & 64.2 & 77.1 & 81.5 & 27.2 & 56.3 & 65.8 & 78.4 & 82.2 \\
    M-Attack & 0.344 & 10.5 & 23.8 & 36.2 & 54.7 & 76.5 & 9.2 & 21.4 & 33.8 & 51.7 & 74.2 & 11.1 & 24.8 & 35.6 & 53.4 & 75.7 \\
    GeoShield & 0.267 & \underline{3.1} & \underline{6.8} & \underline{15.2} & \underline{31.8} & \textbf{46.2} & \underline{2.5} & \underline{5.1} & \underline{13.8} & \textbf{27.4} & \textbf{49.8} & \underline{2.8} & \underline{5.5} & \underline{14.5} & \textbf{26.3} & \textbf{48.7} \\
    Vanilla Edit & \underline{0.893} & 18.3 & 36.8 & 49.0 & 70.0 & 81.4 & 16.5 & 33.2 & 45.2 & 66.4 & 79.8 & 17.1 & 34.5 & 46.8 & 67.5 & 80.5 \\
    \midrule
    \textbf{GeoEdit (Ours)} & \textbf{0.944} & \textbf{2.6} & \textbf{5.7} & \textbf{14.9} & \textbf{30.5} & \underline{53.8} & \textbf{2.1} & \textbf{4.2} & \textbf{13.0} & \underline{28.6} & \underline{51.3} & \textbf{2.3} & \textbf{4.8} & \textbf{13.9} & \underline{29.9} & \underline{52.5} \\
    \textbf{GeoEdit w/o $\mathcal{K}$} & 0.925 & 5.2 & 10.8 & 21.6 & 41.7 & 64.4 & 4.3 & 9.8 & 18.2 & 38.0 & 61.9 & 4.9 & 10.7 & 19.4 & 39.6 & 63.8 \\
    \midrule
    \midrule
    \multirow{2}{*}{\textbf{Method}} & \multirow{2}{*}{\textbf{Fid. ($\uparrow$)}} & \multicolumn{5}{c|}{\textbf{Gemini-2.5-Pro Acc@Km ($\downarrow$)}} & \multicolumn{5}{c|}{\textbf{GLM-4.6V Acc@Km ($\downarrow$)}} & \multicolumn{5}{c}{\textbf{Doubao-Seed-2.0-Pro Acc@Km ($\downarrow$)}} \\
    & & 1km & 25km & 200km & 750km & 2500km & 1km & 25km & 200km & 750km & 2500km & 1km & 25km & 200km & 750km & 2500km \\
    \midrule
    Unprotected & 1.000 & 24.2 & 48.7 & 58.6 & 72.4 & 84.6 & 20.5 & 37.4 & 51.2 & 67.1 & 83.7 & 19.7 & 36.2 & 49.8 & 65.7 & 82.8 \\
    M-Attack & 0.344 & 9.8 & 20.5 & 34.3 & 51.5 & 72.8 & 8.1 & 18.8 & 31.7 & 48.4 & 70.6 & 9.2 & 19.4 & 32.3 & 49.2 & 72.5 \\
    GeoShield & 0.267 & \underline{4.2} & \underline{8.5} & \underline{18.4} & \textbf{31.2} & \textbf{49.5} & \underline{3.1} & \underline{6.5} & \underline{14.8} & \underline{28.5} & \underline{52.8} & \underline{3.0} & \underline{5.8} & \underline{16.2} & \underline{31.5} & \underline{48.6} \\
    Vanilla Edit & \underline{0.893} & 14.6 & 30.8 & 43.2 & 62.5 & 78.2 & 11.4 & 25.1 & 39.0 & 54.2 & 71.5 & 11.0 & 24.8 & 38.9 & 54.1 & 71.3 \\
    \midrule
    \textbf{GeoEdit (Ours)} & \textbf{0.944} & \textbf{3.5} & \textbf{7.9} & \textbf{17.8} & \underline{34.0} & \underline{58.8} & \textbf{2.8} & \textbf{5.9} & \textbf{13.5} & \textbf{26.1} & \textbf{49.0} & \textbf{2.2} & \textbf{4.7} & \textbf{14.9} & \textbf{29.0} & \textbf{46.1} \\
    \textbf{GeoEdit w/o $\mathcal{K}$} & 0.925 & 4.9 & 11.8 & 20.5 & 38.2 & 62.1 & 4.2 & 9.3 & 18.4 & 35.7 & 59.2 & 3.9 & 8.6 & 17.3 & 34.5 & 58.0 \\
    \bottomrule
  \end{tabular}
  }
\end{table*}

\section{Experiments}
\label{sec:experiments}
\subsection{Experimental Setup}

\paragraph{\textbf{Datasets}}
Existing studies~\cite{liu2025geoshield,zhang2025disrupting} often treat geoprivacy protection as an adversarial perturbation problem evaluated on standard geolocation benchmarks such as Im2GPS3k~\cite{im2gps} and YFCC4k~\cite{yfcc}. These benchmarks mainly contain public and benign scenes, such as tourist attractions or landmark check-ins, where location disclosure is intentional. This leads to a mismatch between current evaluation settings and the real-world geoprivacy risk scenarios.

To better reflect real-world privacy risks, we establish \textbf{LifePhotoBench}, a curated subset comprising 1,553 challenging images (383 from Im2GPS and 1,170 from YFCC).
The selection criteria strictly focus on unintentional privacy leakage in everyday scenarios:
\ding{182} Event-centered daily-life photos, where the primary narrative focuses on activities (e.g., dining, walking, resting).
\ding{183} Subject-centered environmental portraits, where a person is the visual focus.
In both cases, explicit landmarks are intentionally excluded. The scene context is not the main focus of the image but often plays a key role in enabling geographic inference through implicit cues. 


\paragraph{\textbf{Evaluation Metrics}}
Geolocation accuracy is measured by the Haversine distance between predicted and ground-truth coordinates, evaluated at five granularities: street (1\,km), city (25\,km), region (200\,km), country (750\,km), and continent (2500\,km). We report the Median Distance Error (MDE) to represent the overall performance.
To assess the {Fidelity (Fid.)} of the protected images, we measure the cosine similarity between the CLIP (ViT-B/32) image embeddings of the original and edited images. This metric evaluates the extent to which the core visual narrative and overall semantic content are preserved after protection.

\paragraph{\textbf{Baselines \& Target Models}}
We compare GeoEdit against: \textbf{Unprotected} original images, \textbf{M-Attack}~\cite{li2025frustratingly} (we use its untargeted variant), \textbf{GeoShield}~\cite{liu2025geoshield}, and a simple \textbf{Vanilla Edit} baseline that directly invokes an MLRM to generate and execute editing prompts without our risk attribution or adversarial verification loop. We adopt the baselines' recommended settings in our implementation to ensure a fair comparison.
To comprehensively evaluate the protection effectiveness, we conduct geolocation attacks using 6 mainstream MLRMs with the adversarial prompt from GeoShield \cite{liu2025geoshield}: {GPT-5.1}, {Claude-4.5-Sonnet}, {Claude-3.7-Sonnet}, {Gemini-2.5-Pro}, {GLM-4.6V}, and {Doubao-Seed-2.0-Pro}.

\paragraph{\textbf{Implementation Details}}
GeoEdit is driven by {\small {Qwen3.5-Plus}} and uses {\small {Doubao-Seedream-5.0-lite}} as the editing model. The fidelity threshold for image editing is set to $\tau=0.25$. The maximum number of ReAct iterations is 15. For the joint reward, we set $w_D=1.0$, $(\tau_N,\tau_R,\tau_C)=(0.5,0.5,0.5)$, $E_{\star}=2000\,\text{km}$, $D_{\star}=500\,\text{km}$, $\alpha=0.85$, $(\lambda_1,\lambda_2,\lambda_3)=(0.01,0.02,0.05)$, and $\lambda_S=0.4$. For insight distillation, we construct a 100-image subset by sampling about 1\% of LifePhotoBench (16 images) and adding other images from Im2GPS3k and YFCC4k. Each sample is explored with $G=3$ outputs for 3 epochs at temperature 0.3, adjusting only one insight at a time.


\subsection{Main Results}

\paragraph{\textbf{Protection Effectiveness}}
Table~\ref{tab:track2} demonstrates that GeoEdit achieves an optimal balance between visual fidelity and privacy protection. 
While visual perturbation baselines (M-Attack, GeoShield) reduce geolocation accuracy, mostly achieving their best confusion at the continent level (2500\,km), they severely degrade image semantics at the cost of dropping fidelity to below 0.35.
In contrast, GeoEdit provides effective protection at fine-grained scales (1km $\rightarrow$ 200km) and transfers consistently across 6 mainstream MLRMs, all while maintaining a high fidelity of 0.944.
Furthermore, the evaluation of the GeoEdit w/o $\mathcal{K}$ variant indicates the importance of the accumulated protection insight for de-evidencing image editing.

\begin{figure}[t]
    \centering
    \includegraphics[width=\linewidth]{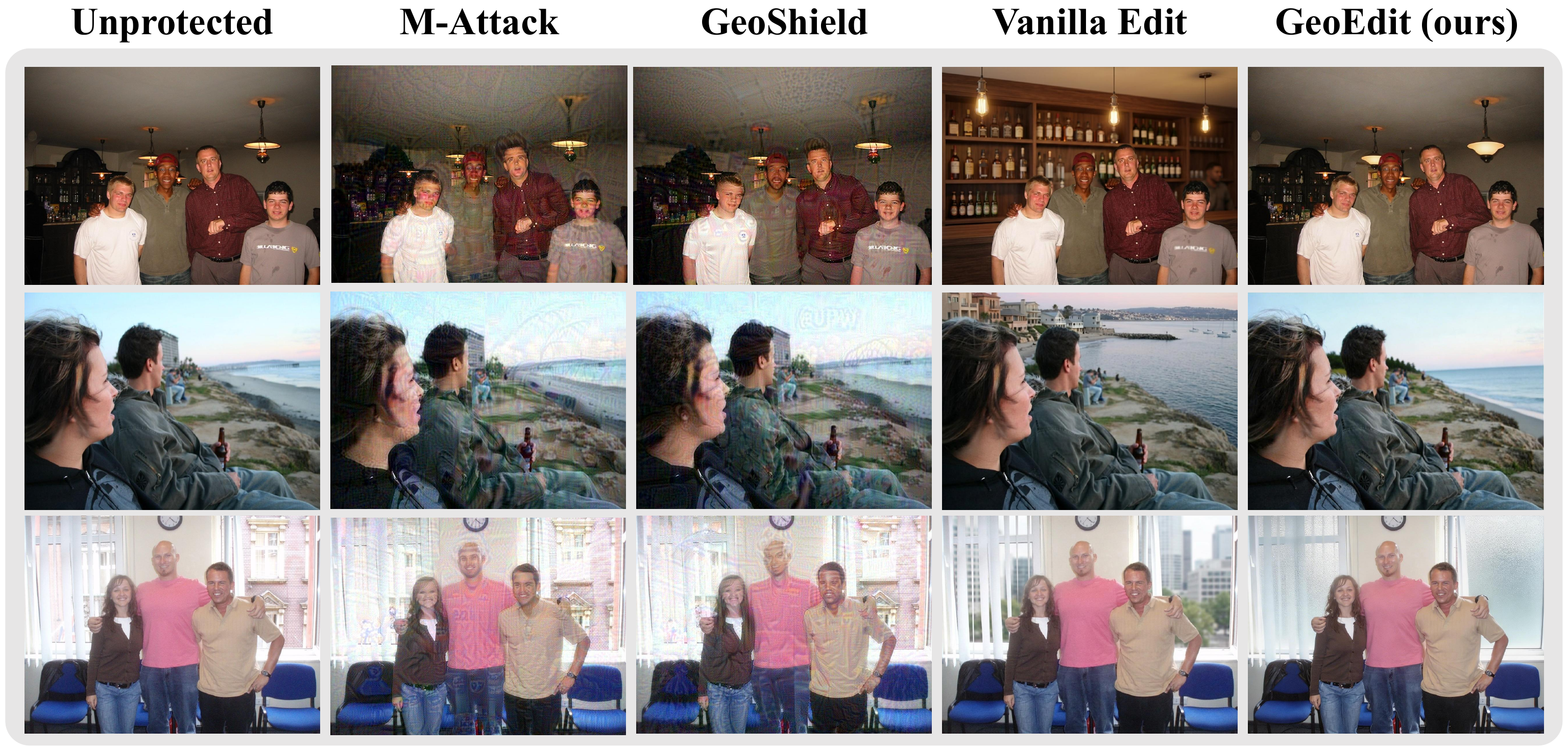}
    \caption{\textbf{Visualization of different protected images.} GeoEdit achieves silent protection while preserving naturalness.}
    \label{fig:exp_cp}
    \vspace{-10pt}
\end{figure}

\begin{table*}[t]
  \caption{Results of visual utility evaluation. W-D-L (\%) is the percentage of Wins, Draws, and Losses for the images under each corresponding setting in blind pairwise A/B tests. Cap-Fid is MLRM's caption fidelity of the protected images.}
  \label{tab:utility}
  \vspace{-5pt}
  \centering
  \resizebox{0.92\linewidth}{!}{
  \begin{tabular}{l|cc|cc|cc|cc}
    \toprule
    \multirow{3}{*}{\textbf{Method}} & \multicolumn{2}{c|}{\textbf{Gemini-2.5}} & \multicolumn{2}{c|}{\textbf{GLM-4.6V}} & \multicolumn{2}{c|}{\textbf{Human 1}} & \multicolumn{2}{c}{\textbf{Human 2}} \\
    \cmidrule(lr){2-9}
    & \multirow{2}{*}{W-D-L (\%)} & \multirow{2}{*}{Cap-Fid $\uparrow$} & \multirow{2}{*}{W-D-L (\%)} & \multirow{2}{*}{Cap-Fid $\uparrow$} & \multicolumn{2}{c|}{W-D-L (\%)} & \multicolumn{2}{c}{W-D-L (\%)} \\
    & & & & & Overall & vs. Orig. & Overall & vs. Orig. \\
    \midrule
    Unprotected {\small (original image) }& 72.0-14.7-13.3 & 1.000 & 84.0-0.0-16.0 & 1.000 & 76.6-8.0-15.4 & 0.0-100.0-0.0 & 83.6-0.9-15.5 & 0.0-100.0-0.0 \\ \midrule
    M-Attack & 25.3-\textbf{25.3}-49.4 & 0.801 & \underline{45.3}-0.0-54.7 & \underline{0.865} & 4.4-\underline{22.8}-72.8 & 0.0-0.0-100.0 & 15.6-2.6-81.8 & 0.0-0.0-100.0 \\
    GeoShield & 10.7-\underline{16.0}-73.3 & 0.798 & 16.0-0.0-84.0 & 0.849 & 5.3-\textbf{24.8}-69.9 & 0.0-0.0-100.0 & 11.0-2.4-86.6 & 0.0-0.0-100.0 \\
    Vanilla Edit & \underline{46.5}-3.5-50.0 & \textbf{0.862} & 33.1-0.0-66.9 & 0.845 & \underline{55.0}-13.1-31.9 & \underline{18.6}-\textbf{16.8}-64.6 & \underline{58.1}-\textbf{2.8}-39.1 & \underline{23.6}-1.6-74.8 \\
    GeoEdit (Ours) & \textbf{62.7}-2.7-34.6 & \underline{0.805} & \textbf{54.7}-0.0-45.3 & \textbf{0.881} & \textbf{67.8}-13.0-19.2 & \textbf{42.7}-\underline{15.4}-41.9 & \textbf{76.0}-\underline{2.5}-21.5 & \textbf{38.2}-\textbf{2.1}-59.7 \\
    \bottomrule
  \end{tabular}
  }
\end{table*}


\paragraph{\textbf{Visual Utility}}
We further evaluate the visual utility of different methods to assess their practical shareability as shown in Table~\ref{tab:utility}.
Specifically, \ding{182} We use \textbf{Caption Fidelity (Cap-Fid, $\uparrow$)} to assess the understanding difference caused by the processing of privacy protection methods. 
We generate captions for both the original and protected images, and calculate the cosine similarity between their text embeddings using Qwen3-Embedding-4B.
\ding{183} We use \textbf{Aesthetic Win-Draw-Loss (W-D-L \%)} to evaluate utility through a pairwise blind A/B test comparing visual appeal and naturalness. 
We utilize MLRMs for large-scale evaluation on LifePhotoBench, and introduce two independent human evaluators for perceptual assessment on 900 pairwise comparisons (approx. 10\% of the full test set) to ensure alignment with human preference. 
Across both automated and human expert evaluations, GeoEdit consistently achieves a highly competitive W-D-L record (e.g., win rates of 62.7\% from Gemini-2.5 and 76.0\% from Human 2), far surpassing M-Attack and GeoShield, which suffer overwhelming losses due to visible artifacts. 
Furthermore, GeoEdit maintains high Cap-Fid scores (0.805 on Gemini-2.5 and 0.881 on GLM-4.6V), confirming that the core narrative remains intact even after de-evidencing.

\begin{table}[t]
  \caption{Ablation results for GeoEdit on geoprivacy protection effectiveness and fidelity on LifePhotoBench. }
  \label{tab:ablation}
  \vspace{-8pt}
  \centering
  \resizebox{\columnwidth}{!}{
  \begin{tabular}{l|c|cccc}
    \toprule
    \multirow{2}{*}{\textbf{Setting}} & \multirow{2}{*}{\textbf{Fid. $\uparrow$}} & \multicolumn{4}{c}{\textbf{MDE (km) $\uparrow$}} \\
    & & {\small \textbf{Gemini-2.5}} & {\small  \textbf{GLM-4.6V}} & {\small \textbf{GPT-5.1}} & {\small \textbf{Claude-4.5}} \\
    \midrule
    \textbf{GeoEdit} & \textbf{0.944} & \textbf{5240.8} & \textbf{5821.3} & \textbf{3645.9} & \textbf{6836.3} \\
    \midrule
    \quad w/o $T_{risk}$ & 0.915 & 2453.6 & 2847.2 & 1954.8 & 3158.4 \\
   \quad w/o $T_{edit}$ & 0.931 & 3582.4 & 4251.6 & 2849.5 & 4526.8 \\
    \quad w/o $T_{eval}$ & 0.898 & 1627.5 & 1853.8 & 1482.2 & 2084.6 \\
    \midrule
    Vanilla Edit & 0.893 & 1354.2 & 1428.5 & 1215.6 & 1583.4 \\
    \bottomrule
  \end{tabular}
  }
\end{table}

\subsection{Further Analysis}

\paragraph{\textbf{Ablation Studies}}
We conduct experiments with three ablation settings: \textbf{w/o $T_{risk}$} removes the Bayesian risk calculation, meaning the core agent selects visual cues directly based on MLRM's internal prior without probabilistic structural guidance; \textbf{w/o $T_{edit}$} bypasses the specialized insight-guided editing synthesis, meaning the core agent has to invoke the image editing tool with its crafted instructions; \textbf{w/o $T_{eval}$} disables the adversarial verification loop, meaning the core agent can only rely on general auxiliary tools to obtain feedback.
We also introduce \textbf{Vanilla Edit} for comparison.

The ablation results indicate that: \ding{182} GeoEdit's advantage comes from the coordinated interaction of risk attribution, insight-guided editing, and adversarial verification, rather than from generic image editing alone. Removing any single module noticeably degrades both privacy protection and visual utility. \ding{183} Notably, disabling $T_{eval}$ causes the most significant drop (e.g., MDE drops from 5821.3\,km to 1853.8\,km on GLM-4.6V), indicating that closed-loop adversarial feedback is the most critical component for effective and reliable geoprivacy protection.


\begin{figure}[t]
  \centering
  \includegraphics[width=0.8\linewidth]{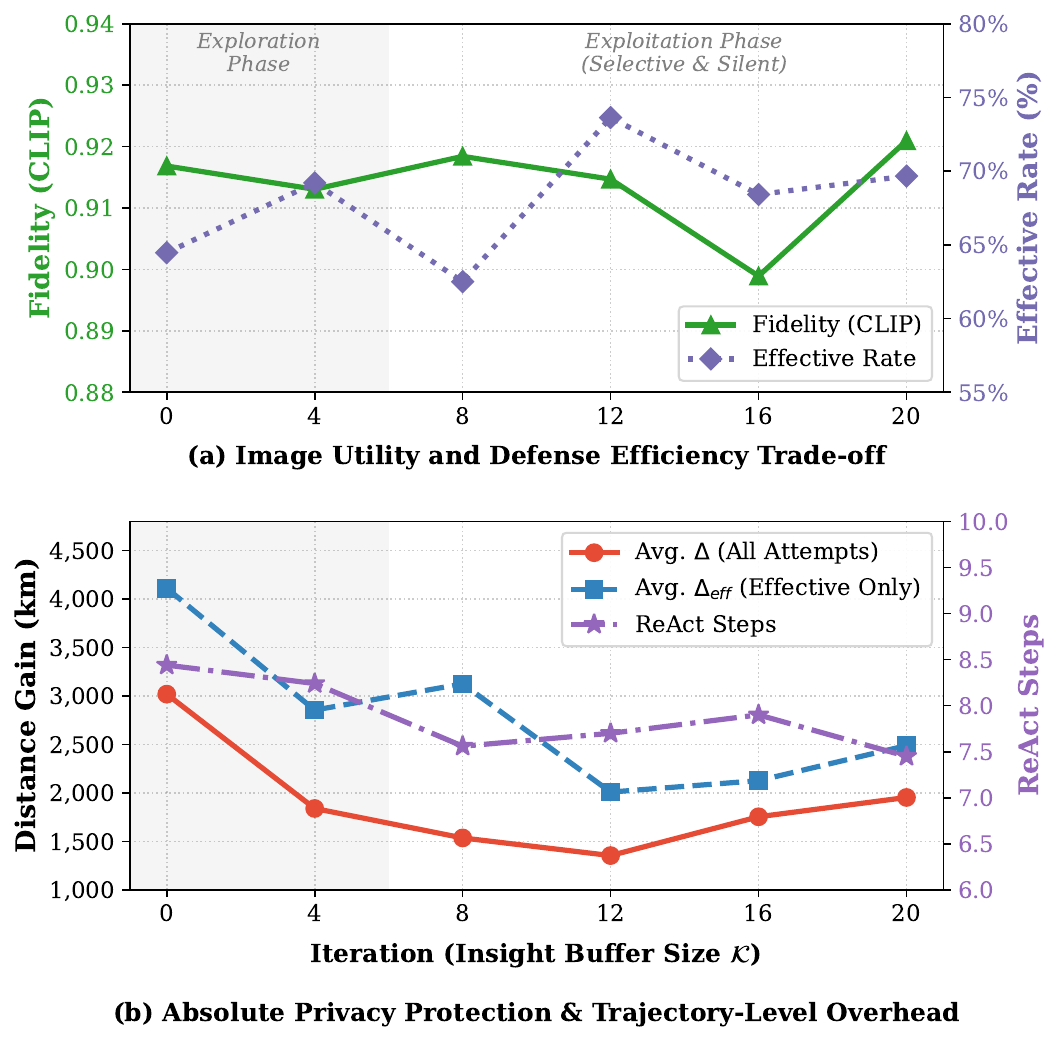}
  \vspace{-8pt}
  \caption{Impact of Protection Insight on GeoEdit. We report the average performance of all edited image attempts and corresponding ReAct steps per trajectory at each insight buffer size during self-evolution.}
  \label{fig:experience_ablation}
  \vspace{-8pt}
\end{figure}


\begin{table*}[t]
  \caption{Interpretability analysis of key protection insights distilled via self-evolution.}
  \label{tab:insights}
  \vspace{-8pt}
  \centering
  \resizebox{0.85\linewidth}{!}{
  \begin{tabular}{m{0.43\textwidth} m{0.18\textwidth} m{0.29\textwidth}}
    \toprule
    \textbf{Example} & \textbf{Editing Insight} & \textbf{Detail of Insight} \\
    \midrule
    \centering\arraybackslash\includegraphics[width=0.37\textwidth]{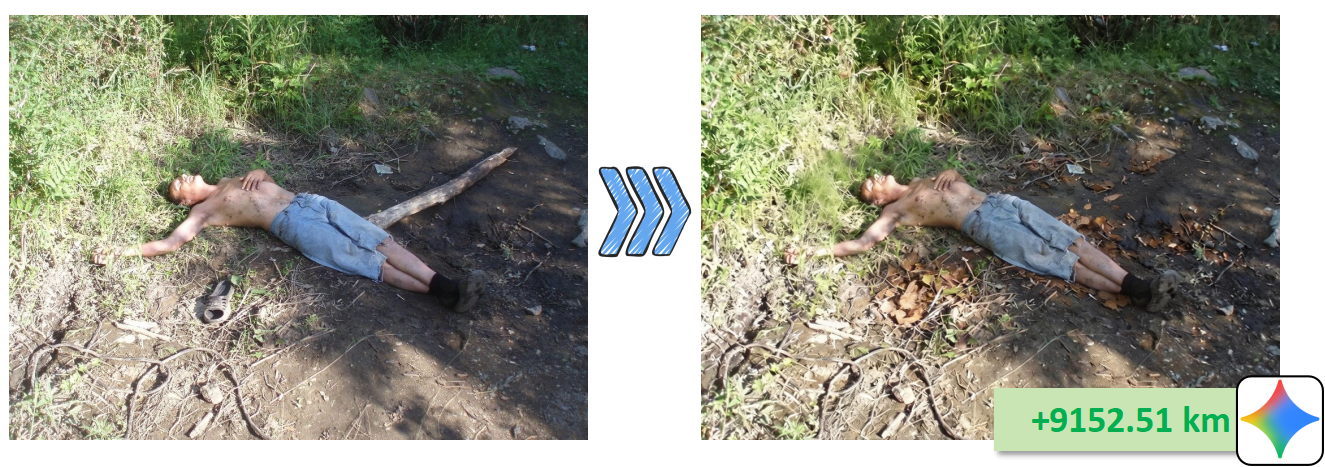} 
    &
    \textbf{Subtractive   \newline Micro-Feature \newline  Removal}
    &
    For non-landmark scenes, prioritize subtractive micro-feature removal (e.g., erasing botanical markers, scattered debris) over additive effects to avoid creating new regional archetypes.
    \\
    \midrule
    \centering\arraybackslash\includegraphics[width=0.37\textwidth]{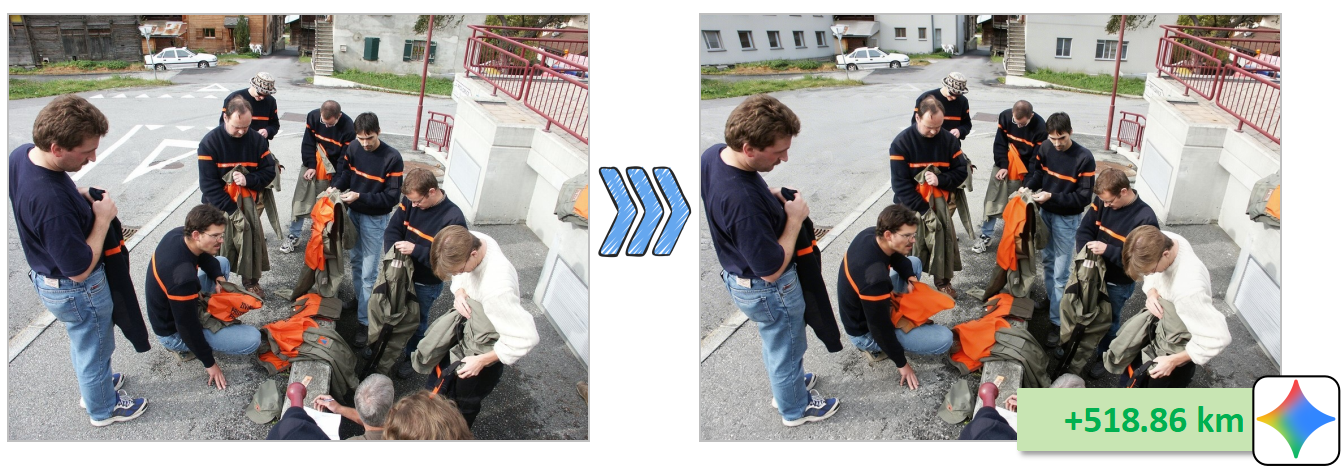} 
    &
    \textbf{Cultural Marker  \newline Neutralization}
    &
    Identify and replace culturally distinctive elements with contextually plausible but globally generic alternatives to create semantic ambiguity while preserving the overall scene narrative.
    \\
    \midrule
    \centering\arraybackslash\includegraphics[width=0.37\textwidth]{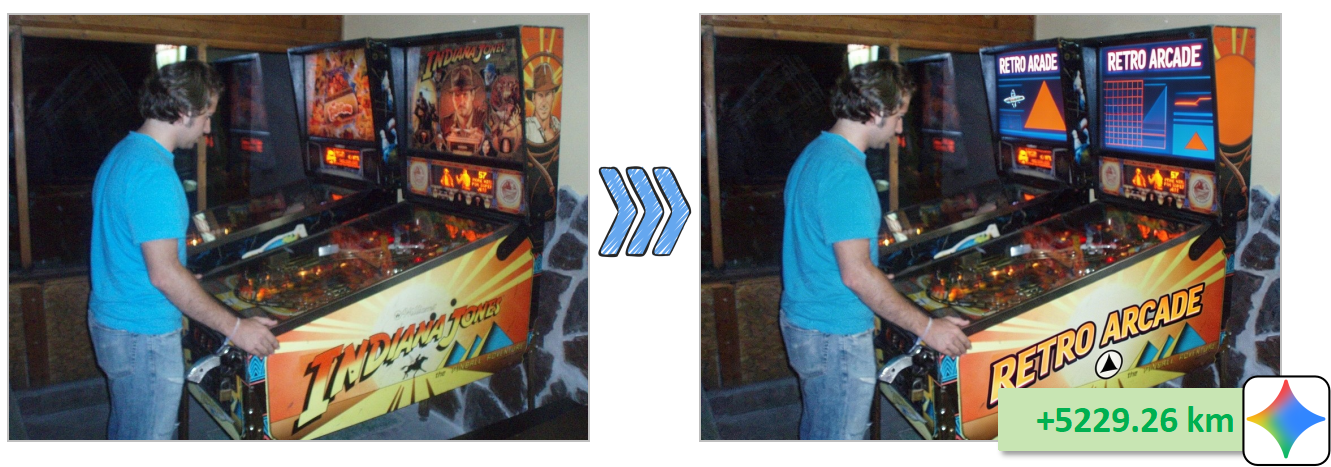} 
    &
    \textbf{Semantic  \newline Simplification}
    &
    Replace distinctive features with neutral, context-appropriate surfaces rather than generating new complex patterns, while maintaining natural weathering and texture noise, specifically optimizing for indoor scenes to maximize ambiguity while preserving naturalness.
    \\
    \midrule
    \centering\arraybackslash\includegraphics[width=0.37\textwidth]{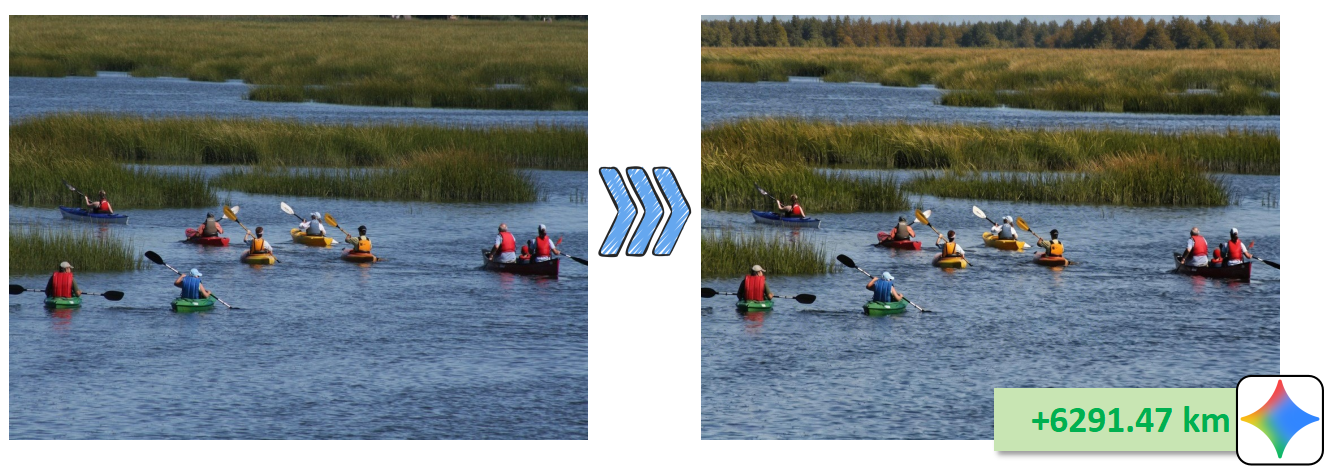} 
    &
    \textbf{Structural \newline Naturalization}
    &
    Replace specific man-made structures with generic natural elements via subtle substitution to break database links and hierarchical narratives while maintaining visual plausibility.
    \\
    \bottomrule
  \end{tabular}
  }
\end{table*}

\paragraph{\textbf{Impact of the Editing Insights}}
To evaluate the value of these insights, we observe the performance changes of GeoEdit during its self-evolution process from the perspective of insight buffer ($\mathcal{K}$) size growth. We introduce the following metrics for analysis:
\begin{itemize}[leftmargin=1em]
    \item \textbf{$\Delta_{\text{all}}$} is the average distance gain (in km), reflecting the total geographical displacement of all edit attempts.
    \item \textbf{$\Delta_{\text{eff}}$} is the average effective distance gain (in km), considering only successful edits that displace geolocation without noticeable artifacts, highlighting the quality of modifications.
    \item \textbf{Eff\%} is the effective edit rate, representing the percentage of successful and meaningful edits (those that cause measurable displacement while maintaining the image's naturalness).
\end{itemize}

As shown in Figure~\ref{fig:experience_ablation}, we illustrate the trade-off between semantic fidelity and defense efficiency (Eff\%). As $|\mathcal{K}|$ increases to 12, the effective rate peaks at 73.6\% while fidelity remains stable around 0.91. Figure~\ref{fig:experience_ablation}(b) shows the absolute privacy protection strength ($\Delta_{\text{all}}$ and $\Delta_{\text{eff}}$) and ReAct Steps. The average distance gain ($\Delta_{\text{all}}$) drops from 3,019 km at $|\mathcal{K}|=0$ and stabilizes around 1,954 km at $|\mathcal{K}|=20$. At $|\mathcal{K}|=0$, the agent executes edits producing large geographical displacements but lower fidelity. As the insight buffer grows, the agent learns to reject edits that degrade visual naturalness, stabilizing around 2,000 km to confuse geolocation at the continent level. More importantly, this transition directly optimizes inference efficiency. 
The accumulation of protection insights significantly accelerates the ReAct convergence, dropping the average iterations from an initial exploratory peak to merely $7.45$ turns per image. 
This highlights the contribution of our self-evolution mechanism in optimizing inference efficiency and mitigating concerns regarding the high computational cost of agentic frameworks.

\begin{figure}[t]
  \centering
  \includegraphics[width=0.8\linewidth]{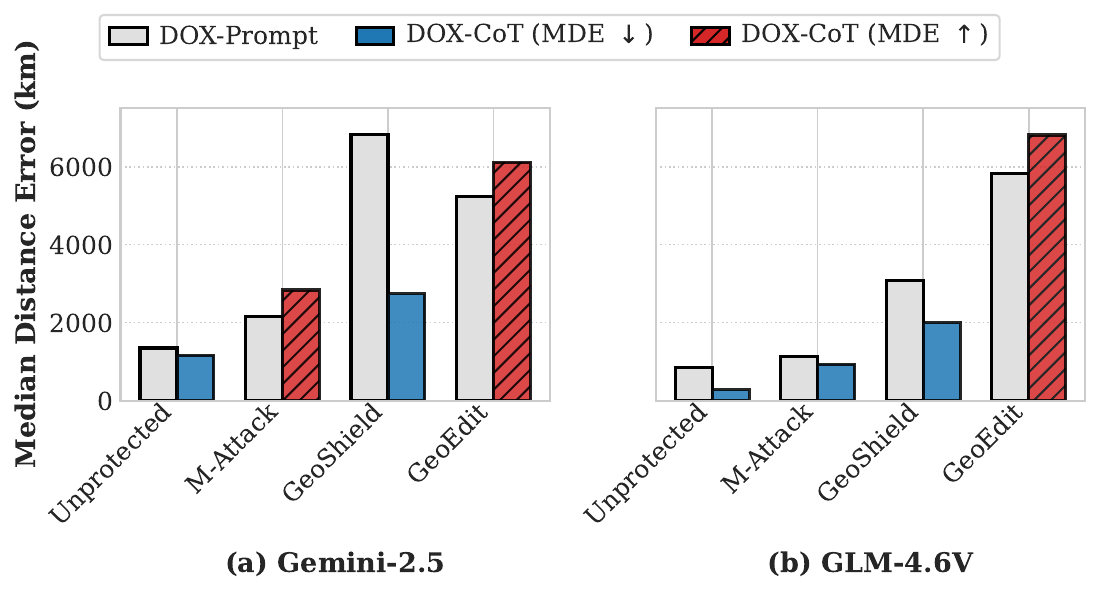}
  \vspace{-5pt}
  \caption{Defense Performance (MDE $\uparrow$, km) under DOX-Prompt and DOX-CoT Probing across two MLRMs. 
  }
  \label{fig:doxcot}
  \vspace{-10pt}
\end{figure}

\paragraph{\textbf{Resistance to Adversarial Dox-CoT Probing}}
To further evaluate the robustness of our defense under reasoning-guided doxing attacks, we employ DOX-COT~\cite{luo2025doxing} for evaluation. As shown in Figure~\ref{fig:doxcot}, DOX-COT reasoning helps the attacker extract and synthesize scattered visual evidence, significantly improving the localization accuracy on unprotected original images.
In contrast, our GeoEdit exhibits a backfire effect, as the semantic edits mislead the reasoning chain. By injecting deceptive context, GeoEdit forces the attacker's CoT to fixate on false evidence, which actually increases the protection distance. For instance, the MDE under Gemini-2.5 increases from 5240.8 km to 6105.2 km, and the GLM-4.6V MDE increases from 5821.3 km to 6813.6 km.
This highlights that our GeoEdit is highly resilient against reasoning-based geographic profiling. 
Furthermore, unlike M-Attack, which relies on uninterpretable adversarial perturbations to achieve backfire, GeoEdit achieves resilience through interpretable semantic modifications.

\begin{table}[t]
\centering
\caption{Robustness of GeoEdit under transformations {\small (Geolocation model: Gemini-2.5)}. Evaluated by Median Distance Error.}
\label{tab:robustness}
\vspace{-10pt}
\resizebox{0.9\linewidth}{!}{
\begin{tabular}{l|ccc}
\toprule
\textbf{Method} & \textbf{Original} & \textbf{JPEG ($Q=50$)} & \textbf{Blur ($\sigma=10$)} \\
\midrule
GeoShield & 6836.3 & 5235.9 & 7198.0 \\
GeoEdit (Ours) & 5240.8 & 5118.1 & 5349.8 \\
\bottomrule
\end{tabular}
}
\vspace{-10pt}
\end{table}

\paragraph{\textbf{Interpretability of Editing Strategies}}
One of the core advantages of GeoEdit is its self-evolution design, which derives explicit geoprivacy protection insights that are both interpretable and reusable.
As shown in Table~\ref{tab:insights}, we present several specific cases where GeoEdit applies the distilled insights to edit challenging images.
For instance, in the first example featuring scattered botanical markers and debris, GeoEdit avoids naive global masking. Instead, guided by the \textit{Subtractive Micro-Feature Removal} insight, GeoEdit explicitly targets and erases these subtle regional archetypes that often serve as critical cues for geolocation models. 
By systematically applying these distilled insights, GeoEdit ensures that its protection mechanism is not only highly interpretable but also provides valuable guidance for human understanding of privacy risks.

\paragraph{\textbf{Robustness to Transformation}}
As shown in Table~\ref{tab:robustness}, GeoEdit's protection remains highly stable under common real-world image transformations. By employing a physical semantic editing approach that modifies the underlying content rather than relying on visual perturbation, GeoEdit ensures consistent stability across different transformations.

\section{Conclusion}
\label{sec:conclusion}


In this paper, we propose \underline{\textbf{Geo}}-Evidence \underline{\textbf{Edit}}ing (\textbf{GeoEdit}), an adversarial de-evidencing framework designed for silent geoprivacy protection via image editing. GeoEdit aims to disrupt the underlying spatial reasoning chains by introducing semantic de-evidencing, thereby achieving an optimal balance between robust privacy protection and visual utility. This de-evidencing process is guided by an interpretable agentic loop that integrates risk assessment, edit planning, and verification. Furthermore, the proposed insight buffer offers interpretable guidance for the editing process, ensuring its transferability across diverse scenarios. 
Evaluated on the LifePhotoBench dataset, GeoEdit achieves a better trade-off in protection, effectively reducing precise geolocation accuracy (Acc@1km) to near 0\% and achieving a Median Distance Error exceeding 5000 km on some models, while maintaining exceptional fidelity of 0.944.

\bibliographystyle{ACM-Reference-Format}
\bibliography{ref}

\end{document}